# Spectral-PQ: A Novel Spectral Sensitivity-Orientated Perceptual Compression Technique for RGB 4:4:4 Video Data

Lee Prangnell and Victor Sanchez
Department of Computer Science, University of Warwick, England, UK

**Abstract** — There exists an intrinsic relationship between the spectral sensitivity of the Human Visual System (HVS) and colour perception; these intertwined phenomena are often overlooked in perceptual compression research. In general, most previously proposed visually lossless compression techniques exploit luminance (luma) masking including luma spatiotemporal masking, luma contrast masking and luma texture/edge masking. The perceptual relevance of color in a picture is often overlooked, which constitutes a gap in the literature. With regard to the spectral sensitivity phenomenon of the HVS, the color channels of raw RGB 4:4:4 data contain significant color-based psychovisual redundancies. These perceptual redundancies can be quantized via color channel-level perceptual quantization. In this paper, we propose a novel spatiotemporal visually lossless coding method named Spectral Perceptual Quantization (Spectral-PQ). With application for RGB 4:4:4 video data, Spectral-PQ exploits HVS spectral sensitivity-related color masking in addition to spatial masking and temporal masking; the proposed method operates at the Coding Block (CB) level and the Prediction Unit (PU) level in the HEVC standard. Spectral-PQ perceptually adjusts the Quantization Step Size (QStep) at the CB level if high variance spatial data in G, B and R CBs is detected and also if high motion vector magnitudes in PUs are detected. Compared with anchor 1 (HEVC HM 16.17 RExt), Spectral-PQ considerably reduces bitrates with a maximum reduction of approximately 81%. The Mean Opinion Score (MOS) in the subjective evaluations show that Spectral-PQ successfully achieves perceptually lossless quality.

**1.0 Introduction**

Due to an increasing consumer demand for a high fidelity visual experience, the utilization of RGB 4:4:4 video data is becoming ubiquitous in various applications including digital cinema, computer vision, machine learning, medical telepathology, home entertainment and video conferencing. Therefore, to attain a high brightness, hue and saturation fidelity experience for the human observer, the direct coding of RGB 4:4:4 data is becoming more prevalent on, for example, High Definition and Ultra HD displays that are capable of the technical capacities specified in ITU-R Recommendation BT.2020 [1] and BT.2100 [2]. This includes High Dynamic Range (HDR) tone mapping, Wide Color Gamut (WCG) and high bit-depth (deep color) RGB 4:4:4 playback. Support for the direct coding of RGB 4:4:4 data is included in the HEVC standard [3, 4] including JCT-VC standardized Range Extensions of HEVC HM (HM RExt) [5] and the Screen Content Coding Extensions of HM RExt (HM RExt + SCM) [6]. This includes the coding of RGB 4:4:4 data of up to 16-bits per sample (i.e., deep color RGB) [1, 2]. Due to the aforementioned HVS spectral sensitivity to photons that are perceived as green, the HEVC standard, by default, treats RGB data in the order of G, B and R. That is, the G channel is treated as the most important perceptual channel; this is similar to the way in which Y is treated as the most important perceptual channel in YCbCr data.

As regards the coding of RGB 4:4:4 video data in HEVC, the raw data is partitioned into Coding Units (CUs), which consist of three equal sized CBs (i.e., a Red CB, a Green CB and a Blue CB) within each CU [7]. Coding techniques, including spatiotemporal prediction [8], transform coding [9], quantization [4, 10, 11] and lossless entropy coding [12], operate in the same manner for both RGB 4:4:4 video data and YCbCr 4:4:4 video data. The main scalar quantization techniques in HEVC, known as URQ [4, 10] and RDOQ [11], are both always enabled by default; however, they are not perceptually optimized. URQ is designed to indiscriminately quantize transform coefficients in G, B and R (or Y, Cb and Cr) Transform Blocks (TBs) at equal levels according to the QStep; the QStep is controlled by a Quantization Parameter (QP) [4, 10]. RDOQ, which is utilized in combination with URQ, is a coefficient-level method designed to quantify quantization-induced distortion and the number of bits required to encode a quantized coefficient. RDOQ chooses an optimal coefficient value, which is subsequently determined by ascertaining an appropriate trade-off between the bitrate and the distortion; this is known as the rate-distortion cost [11].

In general, lossy compression techniques — including transform coding and quantization — are designed to exploit certain psychovisual redundancies including spatiotemporal colour masking and contrast sensitivity [12]. Using the examples of the JPEG standard [13] and HEVC [3, 4], the main redundancy reduction techniques are as follows: Discrete Cosine Transform (DCT)-based integer transform coding [14, 15] and scalar quantization [10, 11]. In JPEG, the DCT basis functions operate according to the Modulation Transfer Function characteristics of the HVS, which, in essence, exploits the Contrast Sensitivity Function (CSF) [12]. After decorrelating an image into the frequency domain, the Direct Current (DC) transform coefficient and the low frequency Alternating Current (AC) transform coefficients contain almost all of the important detail of an image. Therefore, the very high frequency AC coefficients in luma and chroma data can be discarded (zeroed out) by virtue of Quantization Matrices (QMs). The luma QM differs considerably from the chroma Cb and Cr QMs. By default, the luma QM is designed to discard the very high frequency transform coefficients. Conversely, the chrominance QM discards the medium frequency and all high frequency AC coefficients. This is because it is significantly more difficult for the HVS to detect quantization-induced compression artifacts in compressed chrominance data, which means that higher levels of quantization can be applied.

The HEVC standard, in which we implement Spectral-PQ, includes a very similar transform coding and quantization mechanism to the schemes included in the Advanced Video Coding (AVC) standard and the JPEG standard (i.e., DCT and scalar quantization). Focusing on scalar quantization in HEVC and focusing on RGB data, the default quantization techniques are as follows: Uniform Reconstruction Quantization (URQ) [16] and Rate Distortion Optimized Quantization (RDOQ) [11]; they are not perceptually optimized. Neither URQ nor RDOQ take into account the visual masking properties of the HVS; e.g., colour masking in the spatial domain of RGB data. Moreover, URQ and RDOQ cannot distinguish the difference between transformed R, G and B residual data. That is, URQ indiscriminately quantizes R, G and B transform coefficients at equal levels. Similarly, RDOQ manipulates R, G and B quantized coefficients according to Rate Distortion Optimization (RDO) without considering HVS characteristics. Both of these shortcomings typically give rise to coding inefficiency (i.e., the wasting bits by not removing perceptually irrelevant information from the raw RGB data). The HVS is unable to distinguish small differences in shades of colour regardless of the colour space employed, which has led to the development of colour difference metrics including CIEDE2000 [17]. Regarding spectral sensitivity and visual phototransduction, the HVS is most sensitive to photon energies and the associated luminance that humans interpret as green [18, 19] (see Figure 1). Therefore, in the context of image coding and video coding using the RGB colour space as an example, the HVS is the most sensitive to compression artifacts in the Green (G) channel and least sensitive to artifacts in the Blue (B) channel. We exploit this fact in the proposed technique.

To reiterate, in this paper we propose a novel spatiotemporal perceptually adaptive quantization method, named Spectral-PQ. Spectral-PQ exploits HVS psychovisual redundancies inherent in raw RGB 4:4:4 video data. More specifically we employ color masking by accounting for spectral sensitivity of the HVS (i.e., quantizing data in the B and R channels more coarsely), spatial masking in high variance regions of the raw data and PU-level temporal masking in high motion regions of the raw data. Our technique adaptively discards perceptual redundancies in each color channel from the raw RGB 4:4:4 data by virtue of CB-level perceptual quantization. Compared with the aforementioned previously proposed methods, our technique provides some key advantages, which are as follows: 1) Spectral-PQ employs color masking, spatial masking and temporal masking, and 2) Spectral-PQ is adaptive; therefore, every sequence is perceptually compressed according to the unique characteristics of the sequence being processed.

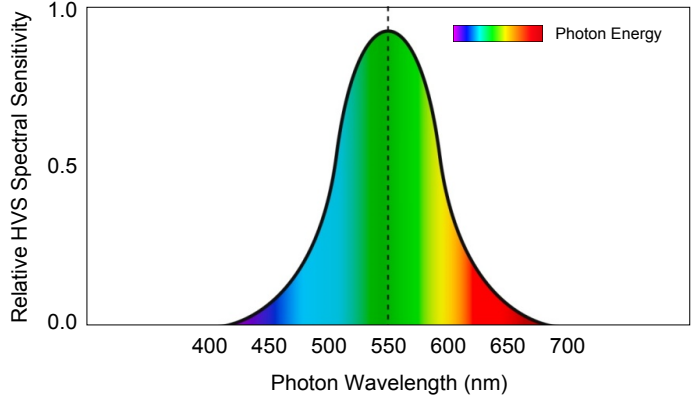

**Figure 1**: A plot showing the relative spectral sensitivity of the HVS to photon energies of various wavelengths (data: National Physical Laboratory, UK). Note how, in terms of the perceived brightness, saturation and hue of colour, the HVS is significantly more sensitive to the photon energies that the HVS interprets as the colour green (495-570 nm).

**2.0 Overview of Visible Light, the Human Visual System and Colour Image Perception**

All forms of electromagnetic radiation in the electromagnetic spectrum, including visible light, microwaves and gamma rays, are different manifestations of light; the photon is the fundamental particle of light. As is the case with all elementary particles, the photon behaves simultaneously as both a particle and a wave (i.e., the wave-particle duality phenomenon in nature, as discovered in the field of quantum mechanics [20]). Visible light is a small range of light in the electromagnetic spectrum that is visible to human observers by virtue of the HVS. The established photon wavelength range in the visible light spectrum is roughly 380 nm to 750 nm, which equates to a frequency range of 668 THz to 484 THz, respectively. Note that the photon energy of visible light ranges from 2 eV to 2.75 eV; therefore, the energy of a photon is inversely proportional to its wavelength. As illustrated in Figure 1 and in terms of spectral sensitivity, the HVS is most sensitive to photons that are perceived as green; note how the plot in Figure 1 corresponds to a Gaussian curve.

Colour vision in humans equates to the combined visual perception of the different photon energies and luminance levels emitted from either natural or synthetic visible light sources. Every aspect of colour that humans visually perceive is ultimately contingent upon the natural processes of visible light (photon energies), physical luminance and the subsequent biological processing of such. In other words, colour is the combined subjective interpretation of electromagnetic radiation in the spectrum of visible light (photon energies) and physical luminance [18, 19].

As previously mentioned, the photon acts as both a wave and a particle. Photon energy $E$ is measured in either J or eV; $E$ is quantified in (1):

$$E = \frac{h \cdot c}{\lambda} \qquad (1)$$

where $h$ and $c$ are established physical constants. Planck's constant $h = 6.626 \times 10^{-18}$ Joule seconds (J), which computes the quantum of action. Constant $c = 3 \times 10^8$ metres per second (m/s) and approximates the speed of light in a vacuum. Lambda $\lambda$ corresponds to the wavelength of a photon in nm. It is important to note that 1 J $= 6.242 \times 10^{18}$ eV.

As an example with relevance to the proposed technique, by utilising the formula in (1), we can quantify the photon energy of electromagnetic radiation — in the visible light range — that humans perceive as red, denoted as $E_{\text{red}}$; $E_{\text{red}}$ is computed in (2).

$$E_{\text{red}} = \frac{(6 \times 10^{-34}\,\text{J} \cdot \text{s}) \cdot (3 \times 10^8\,\text{m/s})}{700 \times 10^{-9}\,\text{m}} = 2.8 \times 10^{-19}\,\text{J} = 1.8\,\text{eV} \qquad (2)$$

Table 1: The wavelength, frequency and energy of photons in the visible light range of the electromagnetic spectrum. This range of photon energies (and the corresponding wavelengths) manifests perceptually as a range of colours in the visual systems of humans, African monkeys, apes and chimpanzees.

| Colour Perception | Wavelength $\lambda$ | Frequency $v$ | Energy $E$ |
|---|---|---|---|
| Violet | 380-450 nm | 668-789 THz | 2.75-3.26 eV |
| Blue | 450-495 nm | 606-668 THz | 2.50-2.75 eV |
| Green | 495-570 nm | 526-606 THz | 2.17-2.50 eV |
| Yellow | 570-590 nm | 508-526 THz | 2.10-2.17 eV |
| Orange | 590-620 nm | 484-508 THz | 2.00-2.10 eV |
| Red | 620-750 nm | 400-484 THz | 1.65-2.00 eV |

As shown in (2), assuming the wavelength of 700 nm, the photon energy of light perceived as red is approximately 1.8 eV (see Table 1). The number of photons emitted per second by a visible light source, denoted as $P$, is contingent upon the energy of the visible light source $L$, measured in J, and also the energy of a photon, denoted as $E$; $P$ is quantified in (3).

$$P = \frac{L}{E} \qquad (3)$$

Photon flux, denoted as $\Phi_P$ is a term used to describe the number of photons emitted per unit area per unit time $U$ (i.e., m$^2$/s). The photon intensity is the photon flux per unit solid angle. The photon flux is computed in (4).

$$\Phi = \frac{P}{U} \qquad (4)$$

The luminous intensity of a visible light source, denoted as $I_v$, is quantified as the wavelength weighted power emitted from the source in a direction per solid angle, which is measured in candela (cd). Luminous intensity corresponds to the intensity of visible light and is, therefore, computed based on the wavelength $\lambda$ of a photon, as shown in (5):

$$I_V = C \cdot V(\lambda) \cdot I_e \qquad (5)$$

where $V(\lambda)$ is the luminous efficiency function, which is standardized by The International Commission on Illumination. $V(\lambda)$ quantifies the average spectral sensitivity of luminance in the human eye. $C$ corresponds to the constant value of 683 lumens per watt (lm/W) and $I_e$ refers to the radiant intensity, which is measured in watts per steradian (W/sr). Note that $V(\lambda)$ is utilized to define the luminous flux, denoted as $\Phi_L$, which is measured in lumens (lm) and is computed in (6):

$$\Phi_L = C \cdot \int_0^\infty V(\lambda) \cdot \Phi_{e,\lambda}(\lambda) \cdot d(\lambda) \qquad (6)$$

where $\Phi_{e,\lambda}$ is the spectral radiant flux, measured in watts per nanometre (W/nm). Recall that variable $\lambda$ corresponds to the wavelength of the photon. Luminance is a measurement of luminous intensity travelling in a given direction; it is measured in candela per square metre cd/m2). From the perspective of light emitting from a light source — such as a TV, a Visual Display Unit (VDU)/monitor, or an incandescent light bulb — luminance computes the light emitted from the light source, which is then distributed within a solid angle. The luminance within a ray of light, denoted as $L_v$, is computed in (7):

$$L_v = n^2 \frac{d\Phi_L}{G} \qquad (7)$$

where $n$ corresponds to the index of refraction of an object, $d\Phi_L$ refers to the luminous flux carried by the beam of the light source and $G$ denotes étendue of a narrow beam containing the ray. The inverse square law must be taken into account because it describes the distribution and the intensity of visible light over arbitrary macroscopic distances. For example, energy that is shown to be twice the distance from the visible light source is spread over four times the area from the source, which equates to the distributed visible light being one fourth the intensity of the visible light in the source. This can be quantified by computing the illuminance $M$ — the amount of luminous flux per unit area — which is quantified in (8):

$$M = \frac{T}{r^2} \qquad (8)$$

where $T$ is the power unit per solid angle, also known as pointance. $T$ is computed in (9):

$$T = \frac{S}{A} \qquad (9)$$

where $S$ is the strength of the visible light source, which can be measured in terms of power, for example, and where $A = 4\pi r^2$, which corresponds to the sphere area of the visible light source.

From the perspective of the Darwinian paradigm of evolutionary biology, the HVS is the product of billions of years of evolution by natural selection [21]. The interaction of photons with the retinal photoreceptor systems facilitates colour vision, whereby the photons are biologically converted into electrical signals in the retina [22]. Visual perception of colour depends on the level of excitation of the different cones. Furthermore, in general terms, the visual cortex system in the brain is responsible for differentiating the signal response received from the Long, Medium and Short (L, M, S) cones. This facilitates the discernment of a vast range of signals that are perceived in the form of a wide range of colours. With a focused concentration on retinal photoreceptors, as shared by all species in the taxonomic order of primate (including humans), rods and cones constitute the key photoreceptors. The retinal photoreceptor system is dominated by rods (120,000,000 units) compared with cones (6,400,000 units). Rods are specialized for low visible light conditions [8, 9]. When subjected to higher intensities of visible light the transmitter release stops because the rod's response to the visible light is much slower than the cone's response. Cones are the retinal photoreceptors that facilitate colour vision and colour perception; they are able to adapt to a vast variety of visible light intensities [18, 19, 22].

In terms of the population of cones, empirical experiments have revealed that 64% are sensitive to photons perceived by the HVS as red, 32% green and 4% blue (i.e., trichromatic colour vision). There are three classifications of retinal cone: L, M, S, each of which contains the transmembrane protein opsin and the molecule chromophore, which are the constituents of photosensitive visual pigments. These pigments are especially sensitive to photons within the following photon wavelength ranges: 650 nm (L), 510 nm (M) and 475 nm (S), which humans interpret as red, green and blue, respectively [18, 19, 22]. To reiterate, although there are more cones that are sensitive to photons which are interpreted as red, the HVS is more sensitive to the perceived brightness of photons that are interpreted by the HVS as green (see Figure 1). In essence, the relationship between visible light and the HVS catalysed the emergence of the Red, Green, Blue (RGB) colour model and the corresponding YCbCr colour space. The Young-Helmholtz theory of trichromatic colour vision is the scientific basis for the RGB colour space. The RGB colour model is an additive tristimulus colour model that is ubiquitous in computer science applications and consumer electronics devices. It amalgamates colour from the following primary colours: red, green and blue, which results in a range of colours depending on the corresponding sample intensity, colour gamut and the associated bit depth. In terms of the physical (hardware) pixels built TVs and monitors, each pixel in these devices contain visible light sources.

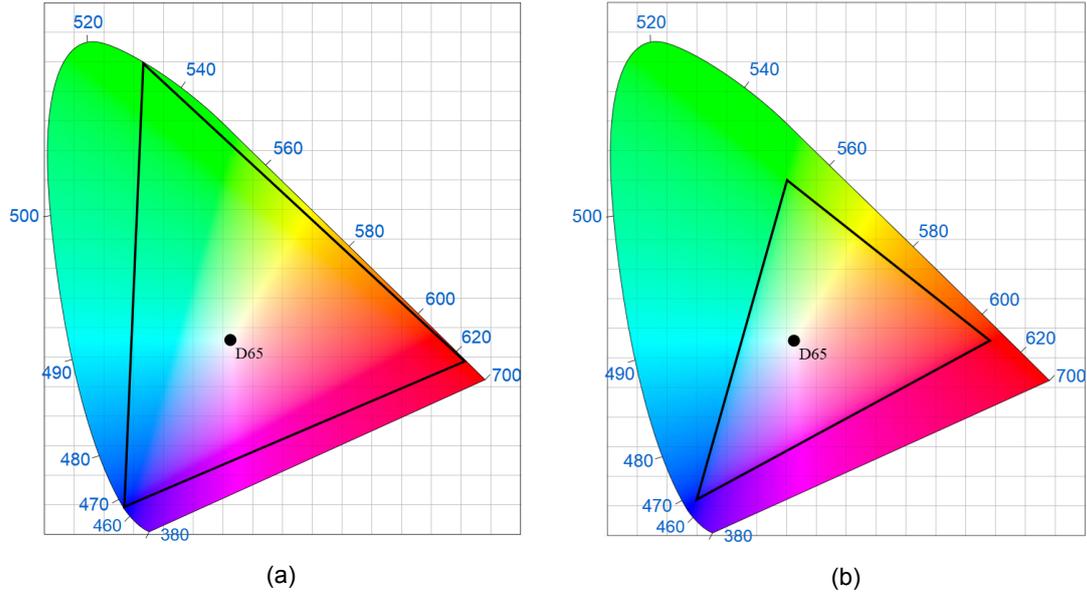

**Figure 2**: Two CIE chromaticity diagrams that highlight standardized ITU-R colour gamuts (i.e., the range of colours inside the triangles), which are relevant to RGB and YCbCr data. Subfigure (a) shows the colour gamut employed in ITU-R Rec. BT.2020 [1] and BT.2100 [2], which are utilized in Ultra HD and HDR applications, respectively. Subfigure (a) shows the colour gamut employed in ITU-R Rec. 709 [23], which is relevant to HD and Standard Dynamic Range (SDR) applications (e.g., HD 1080p 24-bit RGB data). Note that the values in blue refer to the wavelength of photons (in nm).

In a mathematical sense, *Y*, in YCbCr, which is referred to as luma, corresponds to the weighted sum of RGB values (not gamma corrected). The gamma corrected version is denoted as *Y′*. Concentrating on the Hybrid Log Gamma (HLG) version of *Y′* specified in ITU-R Rec. BT.2100 [2] (see Figure 2), *Y′* is defined in (10).

$$Y' = (0.2627 \cdot R') + (0.6780 \cdot G') + (0.0593 \cdot B') \qquad (10)$$

The interaction of the HVS with luminance and photon energies is the physical process by which visible light and the associated luminance is visually perceived as brightness and colourfulness. RGB colour values can be represented by normalized arithmetic, percentage or base-10 integer representations of binary numbers. The binary representations of R, G and B data are dependent on the bit depth of each colour channel. There are $2^t$ sample intensities in each colour channel, where *t* denotes the bit depth of the data. For a bit depth of 8-bits per sample per channel (i.e., 24-bits per sample), the integer value ranges are as follows: $R' \in [0,255]$, $G' \in [0,255]$ and $B' \in [0,255]$. In this example, $R'=0$, $G'=0$ and $B'=0$ represents absolute black (low energy). Conversely, $R'=255$, $G'=255$ and $B'=255$ represents absolute white (high energy). For image or video data with higher bit depths (e.g., 48-bits per sample), this equates to a greater number of colours in each sample. For 48-bit image or video data, the value ranges are as follows: $R' \in [0,65535]$, $G' \in [0,65535]$ and $B' \in [0,65535]$, where $R'=65535$, $G'=65535$ and $B'=65535$ represents absolute white.

The YCbCr colour space is a colour transformation from a given RGB colour space; YCbCr comprises one luma channel (Y) and two chroma channels (Cb and Cr). In YCbCr, luma (Y) refers to an achromatic colour channel that is derived via an approximation of gamma-corrected luminance. The human perception of the brightness of colour in the luma channel is conceptualized as relative luminance, in which the values are normalized to 1 or 100. On a percentage scale, 0% represents absolute black and 100% represents absolute white; moreover, the luma channel contains the vast majority of the finer detail in an image. Chrominance (Cb and Cr) refers to the "difference" colour channels, which are as follows: blue difference (Cb) and red difference (Cr) with reference to the luma (Y) channel; Cb and Cr collectively correspond to the saturation of the colour in an image.

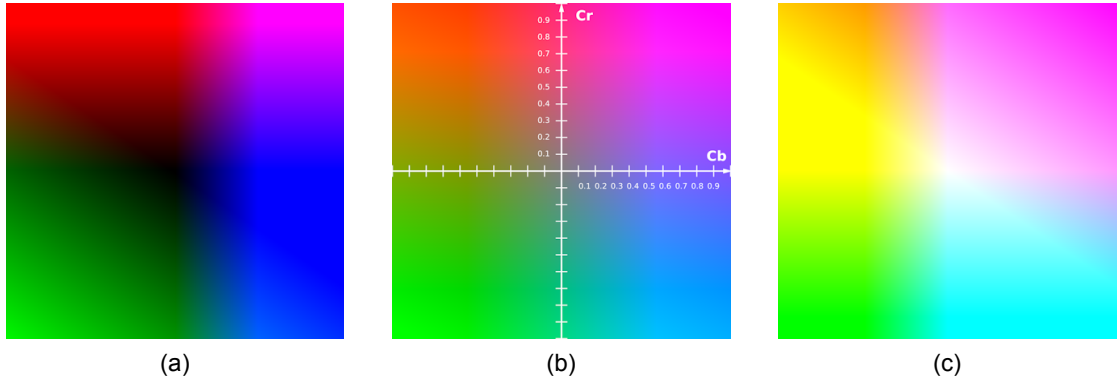

(a)                     (b)                     (c)

**Figure 3**: The chrominance Cb and Cr colour planes. In these examples, the Cb and Cr planes are in the range [−1,1] for a normalized Y value (i.e., $Y \in [0,1]$). Subfigure (a) shows the Cb and Cr colour planes when $Y = 0$; subfigure (b) shows the Cb and Cr colour planes when $Y = 0.5$; subfigure (c) shows the Cb and Cr colour planes when $Y = 1$.

Note that the parameter values applied to the gamma corrected RGB data are weights determined by a luminosity function. It has been shown, in empirical testing, that humans are more sensitive to photons perceived as green in terms of brightness perception, in which case $G'$ is assigned the largest weight. $Cb'$ and $Cr'$ are colour difference channels with reference to the $Y'$ colour channel; the perceived hue and saturation in $Cb'$ and $Cr'$ are dependent on the value of $Y'$ in (10); see Figure 3. $Cb'$ and $Cr'$ are computed in (11) and (12), respectively.

$$Cb' = \frac{B' - Y'}{1.8814} \quad (11)$$

$$Cr' = \frac{R' - Y'}{1.4746} \quad (12)$$

Note that YCbCr 4:4:4, 4:2:2 and 4:2:0 refer to the sampling (resolution) ratios of the chrominance data with respect to the resolution of the luma data. In 4:4:4 data, the Cb and Cr data are the same resolution as the Y data (i.e., no chroma subsampling). The YCbCr 4:2:2 and 4:2:0 versions use spatial chroma subsampling, which is a form of compression. In 4:2:2 data, the Cb and Cr data is half the resolution of the Y data, and in 4:2:0 data, the Cb and Cr data are quarter the resolution of the Y data.

Perceptual considerations have always been a focus of concern during the development of image coding and video coding platforms. H.263 [24], JPEG [13], Moving Picture Experts Group version 4 (MPEG 4) [25], AVC [26, 27] and HEVC [3, 4] all include perceptual compression techniques including the aforementioned integer transform coding and scalar quantization methods. In early image coding research, scientists discovered that spatially subsampling the analogue chrominance components (U and V) in YUV analogue video data is usually imperceptible to the HVS on — now obsolete — Cathode Ray Tube (CRT)-based visual display technologies. This engendered YUV 4:2:0 (analogue) and also YCbCr 4:2:0 (digital) chroma subsampling. Concentrating on the coding of raw digital data, this means that YCbCr 4:2:0 video data, for example, has already been compressed prior to being further compressed by a video coding platform such as HEVC and/or AVC. Although the aforementioned compression techniques, including integer transform coding and scalar quantization, are designed to reduce certain redundancies in raw video data, these methods can be perceptually optimized in order to maximally reduce perceptual redundancies in the image or video data. Furthermore, it is important to note that chroma subsampling compression artifacts in 4:2:0 data are typically conspicuous on contemporary 4:4:4 and HDR capable TVs and Visual Display Units (VDUs).

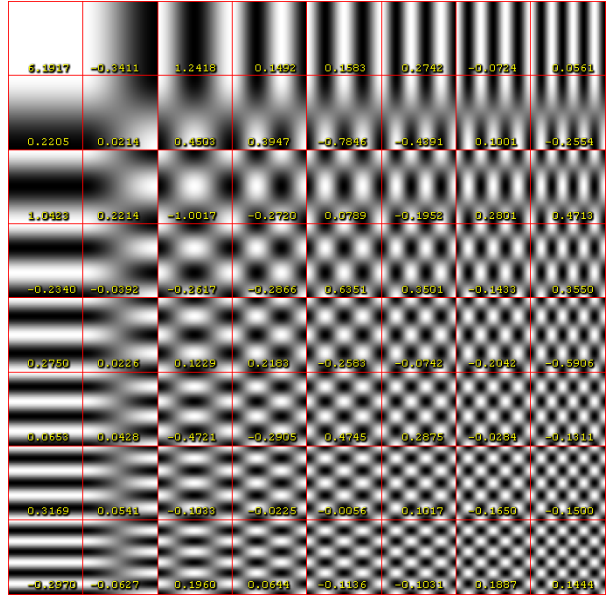

**Figure 4**: Basis functions in the Discrete Cosine Transform version 2 (DCT-II) [14], as utilized in various image and video coding platforms. Note that the numbers in this image correspond to examples of double precision transform coefficients, which are produced after the DCT-II linear transformation has been applied to the original values. In image and video coding standards including JPEG and HEVC, integer approximations of the DCT-II are employed.

In HEVC and JPEG, for example, the DCT-based integer transformation is designed to preserve the energy of the low frequency coefficients. Assuming that the quantization levels applied to the coefficients are not too coarse, the lossy-coded reconstructed picture should contain compression artifacts that are acceptable to the human observer. It is important to note that lossy picture compression is, in actuality, all about how the HVS interprets photon energies and luminance on display units. In the perceptual domain, compression artifacts in a lossy-coded picture centre on a human's visual perception of colour (i.e., brightness, hue, saturation and contrast). Perceptually conspicuous blocking artifacts in lossy-coded data are, in essence, reconstruction errors of the original pixel data. According to the Weber-Fechner law [28, 12], the Just Noticeable Distortion (JND) absolute threshold (the maximum visibility threshold for visual-orientated applications) is defined as the minimum change in a physical stimulus that is perceptible if the corresponding threshold is exceeded. The Weber-Fechner law decrees that there is a mathematical relationship between the subjective sensation of a physical stimulus and the intensity of the actual physical stimulus. This implies that there exists a mathematical relationship between human perceived brightness and the intensity of physical luminance in nature. Likewise, it also implies that there is a mathematical relationship between perceived colour and the energy of photons. The scientific basis of JND facilitates, to a considerable degree, the perceptual compression of picture data. Assuming that the aforementioned visibility threshold is not exceeded, visually lossless quality in the compressed data is successfully achieved; this has been historically confirmed with the utilization of subjective evaluations and an MOS.

Contrast sensitivity is a property of the HVS that is strongly linked to colour perception; it is conceptualized both in terms of achromatic contrast (black and white or greyscale) and chromatic contrast. The discovery of the property of contrast sensitivity of the HVS catalyzed the emergence of CSF/MTF modelling. CSF-based models endeavour to describe how sensitive the HVS is to sine wave gratings as a function of their spatial frequency in cycles/degree; MTF can be conceptualized as a portion of the CSF, whereby the MTF is normalized to 1. Therefore, from the perspective of linear transformations, such as DCT [14], an MTF of 1 corresponds to the Direct Current (DC) transform coefficient, in which most of the important energy resides (see the top-left square in Figure 4; the numerical value is the DC coefficient). Lossy picture coding techniques typically exploit the MTF, whereby CSF-related high spatial frequencies are discarded via linear transform coding and quantization.

| | | | |
|---|---|---|---|
| DC 108 | AC −6 | AC 1 | AC 0 |
| AC 8 | AC 4 | AC 0 | AC 0 |
| AC 2 | AC 0 | AC 0 | AC 0 |
| AC 0 | AC 0 | AC 0 | AC 0 |

**Figure 5**: A toy example of a 4×4 block of quantized transform coefficients. The quantized DC coefficient is shown in the top left (darker orange), the quantized low frequency AC coefficients are shown in lighter orange, the quantized medium frequency AC coefficients are shown in yellow and quantized high frequency AC coefficients are shown in grey.

In terms of linear transforms, these mathematical constructs are heavily utilized in lossy image and video compression systems. DCT and its variants, the Discrete Sine Transform (DST), the Walsh Hadamard Transform (WHT) and the Discrete Wavelet Transform have proved to be extremely useful. The Karhunen-Loève Transform (KLT) is considered to be the so-called gold standard by which most linear transforms are judged (in terms of energy compaction properties and reconstruction of values after the transform's inverse). However, because KLT is image dependent, it is considered by many to be impractical for image and video compression applications. Note that DCT and its variants (DCT-II in particular) are the closest approximations to KLT. This is the reason why DCT is employed as the default transformation in state-of-the-art image and video compression methods, such as HEVC and JPEG. It should be noted, however, that image and video compression standards include integer approximations of the aforementioned linear transforms. This is because the calculations in double precision and floating-point precision transforms, such as DCT, would prove to be much too computationally expensive. Note that the integer approximation of the transform is not invertible in the inverse transform process. In other words, the original values are not fully recoverable due to rounding errors; only close approximations of the original values are recoverable. This is not a problem in lossy compression applications because scalar quantization is typically always employed in combination with transform coding. Scalar quantization is always a lossy process regardless of numerical precision; therefore, the inverse quantization and inverse transformation processes will always incur rounding errors of the original numerical values.

Using HEVC or JPEG as an example, the DCT basis functions operate according to an MTF. After decorrelating an image into the DCT frequency domain, the Direct Current (DC) transform coefficient and the low frequency Alternating Current (AC) transform coefficients typically contain almost all of the important detail of an image. Therefore, the very high frequency AC coefficients in luma and chroma data can be discarded by virtue of quantization. In JPEG, the luma Quantization Matrices (QMs) differ substantially from the chroma Cb and Cr QMs [13]. Higher levels of quantization are applied to chroma image data because quantization-induced artifacts are difficult to detect in compressed chrominance data. As previously implied, the HVS is unable to distinguish small differences in shades of colour regardless of the colour space employed, which has given rise to colour difference methods including CIEDE2000 [17]. To reiterate, the HVS is most sensitive to photon energies — and the associated luminance — that are perceived as green [18, 19] (see Figure 1). In the context of image and video coding using the RGB and YCbCr colour spaces examples, the HVS is the most sensitive to quantization-induced compression artifacts in the G or Y channel and least sensitive to artifacts in the B or Cb channel. In essence, spectral sensitivity appears to have been considered somewhat in the JPEG standard; however, subsequent coding standards, including AVC/H.264 and HEVC, appear to have abandoned quantizing chroma data differently from luma data (by default).

Table 1: The first six values of *QP*, *QStep*, *m* and *s*.

| QP | 0 | 1 | 2 | 3 | 4 | 5 |
|---|---|---|---|---|---|---|
| QStep | 0.6300 | 0.7071 | 0.7937 | 0.8909 | 1.0000 | 1.1225 |
| m | 26214 | 23302 | 20560 | 18396 | 16384 | 14564 |
| s | 40 | 45 | 51 | 57 | 64 | 72 |

**3.0 Overview of Relevant Methods and Techniques**

Because Spectral-PQ is integrated into the HEVC standard, it is appropriate to provide a detailed background in terms of how HEVC operates at the level of transform coding and quantization. JCT-VC developed and standardized HEVC to supersede the presently ubiquitous Advanced Video Coding (AVC)/H.264 standard. The standardization of HEVC version 1 took place in January 2013; note that HEVC version 5 is the latest version of the standard. In comparison with AVC, the key improvement that HEVC attains is the outstanding coding efficiency improvement that it yields. With its enhanced video coding algorithms, HEVC improves coding efficiency by up to 50% compared with AVC. These vast improvements in coding efficiency facilitate the coding and decoding of high quality bitstreams for utilization on appreciably high display resolution environments (including Ultra HD 4K and 8K).

HEVC includes finite precision integer approximations of the DCT and the DST [9]. These techniques transform intra prediction and inter prediction residual data from the spatiotemporal domain into the frequency domain. Recall that the DC transform coefficient and the low frequency AC transform coefficients contain the most important energy in terms of how the HVS perceives the reconstructed video data. As such, after intra prediction and/or inter prediction, DCT and DST are applied to the corresponding residual signals, from which transform coefficients are derived. More specifically, the DCT is applied to intra residual luma and chroma residual blocks of size 8×8 to 32×32. For inter predicted residuals, the corresponding integer approximation of DCT is utilized on 4×4 to 32×32 luma and chroma residual blocks. Note that, for 4×4 intra residue, the DST is utilized instead of DCT. Recall that the integer DCT and DST schemes in HEVC exploit the MTF characteristics of the HVS. This is achieved by compacting the energy of luma and chroma prediction residual samples into the DC coefficient and the very low frequency AC coefficients.

As previously mentioned, it is well known that the DC coefficient and the low frequency AC transform coefficients are more important than the high frequency AC coefficients (in terms of how the reconstructed signal is perceived by the HVS). Because each coefficient frequency sub-band in a TB constitutes a different level of perceptual importance in a compressed picture signal, the distance of AC coefficients from the DC coefficient can be quantified in terms of Euclidean distance. That is, the DC coefficient is the starting point and the distance of each AC coefficient from the DC coefficient represents the perceptual importance of the current AC coefficient.

Recall that URQ is the default uniform quantization method in HEVC. Assuming that chroma QP offsets are not employed in HEVC, the QStep computation for luma data is identical to the QStep computations applied to chroma Cb and Cr data. In terms of the quantization of G/Y, B/Cb and R/Cr transform coefficients using the default URQ and RDOQ combination, the association of the QP and QStep with the Multiplication Factor (MF) and the Scaling Factor (SF), the quantized transform coefficient within all TBs (i.e., all GBR/YCbCr TBs) — denoted as *t* — is computed in (13):

$$t = \frac{X \cdot (m + f)}{2^{21 + \frac{QP}{6} - \log_2 N}} \tag{13}$$

where $X$ denotes the transform coefficient, $m$ corresponds to the MF associated with the QP and $o$ refers to the offset corresponding to the error level incurred by quantization rounding including the level of deadzone; $f = 2^{18}$. Variable $N$ denotes the $N$ value of an $N \times N$ TB. The inverse quantized transform coefficient, denoted as $X'$, is computed in (14):

$$X' = \frac{t \cdot s \cdot 2^{\frac{QP}{6}}}{2^{\log_2 N - 1}} \tag{14}$$

where $s$ is the SF employed for inverse quantization. The URQ method in HEVC is designed such that coefficients in a TB are equally quantized according to the frame level QP; therefore, a single QP value is applied to an entire TB of transform coefficients. MF $m$ and SF $s$ are computed in (15) and (16), respectively.

$$m \approx \left[ \frac{2^{14}}{QStep} \right] \tag{15}$$

$$s = \left[ 2^6 \times QStep \right] \tag{16}$$

Due to the MF and the associated bitwise operations, the values associated with quantization and inverse quantization are quantified without the need for divisions and floating-point operations. Moreover, as shown in Table 1, the MF is inversely proportional to the QP and the QStep. Therefore, decreases to the MF will induce greater levels of quantization. One main objective is to ensure that an increment of $QP$ (i.e., $QP + 1$) equates to an increase of $QStep$ by approximately 12%.

Rate Distortion Optimized Quantization, which is used in combination with URQ, is enabled by default in the JCT-VC HEVC HM software [29]; it is, therefore, the default quantization technique when following the common test conditions. RDOQ is a soft decision quantization method that individually quantizes coefficients in both luma and chroma TBs. This is achieved by minimising the rate-distortion Lagrangian cost function [11]. RDOQ is designed to search for an optimal set of quantized coefficients in order to establish a suitable trade off between bitrate and quantization-induced distortion; as such, a calculation for each transform coefficient is performed separately. In essence, RDOQ manipulates the quantized transform coefficients according to the final RD performance [11]; therefore, it significantly outperforms URQ in terms of reducing bitrates.

In luma and chroma TBs of size $N \times N$, each transform coefficient $X$ with RDOQ is quantized to three level values, which are as follows: 0, $l_1$ and $l_2$. According to [11], for each transform coefficient position in a TB, the Lagrangian cost of each value of $X$ is calculated when the quantization level value, denoted as $l_i$, is equal to 0, $l_1$ or $l_2$. When $X$ is quantized to value $l_i$, the Lagrangian cost $J(\lambda, l_i)$ is computed as follows in (17):

$$J(\lambda, l_i) = r(X, l_i) + \lambda \cdot b(l_i) \tag{17}$$

where $\lambda$ denotes the Lagrangian multiplier (the value is computed in [6], where $r$ denotes the quantization error if coefficient $X$ is quantized to level $l_i$ and where $b$ corresponds to the number of bits required to code $l_i$. Variables $l_1$ and $l_2$ are computed in (18) and (19), respectively.

$$l_1 = \left\lfloor |X| \cdot \frac{m}{2^{15+\frac{QP}{6}}} \right\rfloor \tag{18}$$

$$l_2 = l_1 + 1 \tag{19}$$

Recall that $m$ is the MF, as computed in (15). The final quantized level, denoted as $q$, is computed in (20). Therefore, the Lagrangian cost function is updated to $J(\lambda,q)$.

$$q = \arg\min J(\lambda, l_i) \tag{20}$$

A notable drawback of RDOQ is the computational complexity associated with the rate-distortion decisions it carries out. Recall that it is designed primarily to select an optimal quantization level, with (17), to find a suitable trade off between rate and distortion; this process alone requires significant computational complexity [30].

3.1 Review of Contemporary RGB 4:4:4 Video Coding Techniques

In terms of the relevant state-of-the-art algorithms that are present in the literature, Song et al. [31] propose a block adaptive inter-color compensation scheme for RGB 4:4:4 video coding. It reduces inter-color redundancy in RGB channels by employing a novel linear model; this method achieves coding efficiency gains by approximately 20%. Kim et al. [32] propose an inter-color redundancy technique for RGB 4:4:4 video coding in which they utilize an adaptive inter-plane weighted prediction method; this method attains coding efficiency gains of up to 21%. Zhao and Ai [33] propose a color redundancy reduction technique for RGB 4:4:4 intra coding. In this technique, the authors employ an adaptive inter-color prediction scheme whereby the B and R components are predicted from the G component; this method improves coding efficiency by around 30%.

Huang et al. [34] propose an adaptive weighted distortion scheme for utilization in the Rate-Distortion Optimization (RDO) process within HEVC. In this method, PSNR is quantified by taking the importance of the G channel — relative to the B and R channels — into account; this technique improves coding efficiency by up to 44%. Huang and Lei [35] propose a cross component technique designed for the inter-prediction process in HEVC. In this method, statistical correlations within the R, G and B color components are identified in the motion compensation prediction signal, from which a residual prediction parameter is derived; this technique achieves coding efficiency gains of up to 31%. Shang et al. [36] propose a perceptual quantization technique for HEVC. In this method, the authors develop frequency-weighting matrices based on a Contrast Sensitivity Function (CSF) model, from which QMs are derived. This method is a spatial masking technique that operates at the transform coefficient level; it achieves coding efficiency gains of up to 21%.

The principle drawback of the previously proposed RGB 4:4:4 video coding techniques proposed in [31]-[36] is that they are not perceptually optimized. Though they achieve noteworthy coding performance improvements, the bitrate reductions could have been much greater had they taken into account the spatiotemporal perceptual redundancies inherent in the raw RGB 4:4:4 data. The QM technique proposed in [36] accounts for HVS perceptual redundancies; however, the method has certain shortcomings. The authors of this method adopt a spatial CSF model that is not designed for RGB 4:4:4 data. Furthermore, this QM technique comprises static, non-adaptive intra and inter QMs for spatial masking only. This method does not account for the characteristics of different RGB 4:4:4 video sequences, nor does it take temporal information into account.

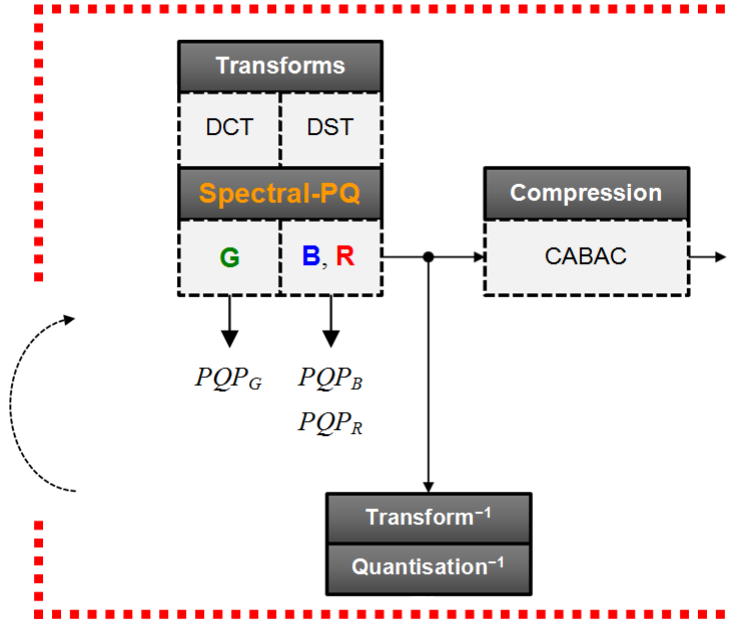

**Figure 6**: A block diagram that shows a graphical representation of the proposed Spectral-PQ method employed with HEVC HM 16.17. The red dotted line indicates the area within the HEVC coding pipeline in which Spectral-PQ operates. Variables $PQP_G$, $PQP_B$ and $PQP_R$ denote the perceptual QPs at which visually lossless compression is achieved.

**4.0 Proposed Spectral-PQ Algorithm**

Due to the increasing demand for direct RGB data coding, it is desirable to develop perceptual coding algorithms that cater for raw RGB data. Pertaining to video coding, the direct compression of RGB data of various bit depths is a recent feature; it did not exist prior to the Format Range Extensions (RExt) release of the HEVC standard. However, this feature is not perceptually optimized in HEVC RExt and therefore leaves room for improvement.

With the proposed Spectral-PQ technique, raw data in the RGB colour space can be compressed to a substantial level without incurring a perceptible loss of visual quality in the reconstructed data. According to the visible light-based and HVS spectral sensitivity model intrinsic to Spectral-PQ, the proposed method is employed with HEVC. Spectral-PQ quantizes data in G, B and R CBs to maximum levels (from the perspective of perceptual coding), thus ensuring that the compressed data is visually indistinguishable from the raw data. As previously mentioned, the perceptual compression is achieved by virtue of colour spatial masking based on the aforementioned visible light and spectral sensitivity model. As a by-product of this mechanism, Spectral-PQ also exploits the MTF-based transform coding design inherent in HEVC. To reiterate, the primary objective of Spectral-PQ is to accomplish visually lossless coding in accordance with trichromatic colour vision theory.

Spectral-PQ is a spatiotemporal perceptual quantization technique in which HVS spectral sensitivity, spatial variance and motion vector information are exploited. Spectral-PQ is based on similar principles to our previously proposed methods known as C-BAQ [37] and FCPQ [38]. Both C-BAQ and FCPQ are spatial-only perceptual quantization techniques designed to improve upon AdaptiveQP [39]. AdaptiveQP is a perceptual quantization method that works in combination with URQ and RDOQ in HEVC to further reduce bitrates. It is designed to compute the spatial variance of G (or Y) CBs only. Consequently, AdaptiveQP adjusts the QP of an entire $2N \times 2N$ CU without taking into account the variance of data in the sub-blocks of chroma Cb (or B) and Cr (or R) CBs, which is a significant shortcoming of the technique.

With application for the perceptual coding of YCbCr data, C-BAQ [37] improves upon AdaptiveQP by taking into account the combined spatial variance of data in the sub-blocks of Y, Cb and Cr CBs; the CU-level QP is thus adjusted accordingly. FCPQ [38] expands upon C-BAQ by computing QPs at the CB level. That is, FCPQ separately adjusts the QPs for the Y CB, the Cb CB and the Cr CB based on the variances of data in the sub-blocks of all three CBs. Consequently, three CB-level QPs are signaled in the Picture Parameter Set (PPS) [40, 41]; this is also the case for Spectral-PQ. Spectral-PQ significantly improves upon AdaptiveQP, C-BAQ and FCPQ in the following areas: Spectral-PQ accounts for color masking, spatial masking and temporal masking. Furthermore, AdaptiveQP, C-BAQ and FCPQ are designed to perceptually decrease the QP for smooth (low variance) regions of raw data regardless of the QStep employed during the coding process, which unnecessarily increases bitrates. Because Spectral-PQ employs color masking and temporal masking, it is not necessary to decrease the CB-level QP in smooth regions of an RGB 4:4:4 sequence.

In terms of the Spectral-PQ algorithm, the CB-level perceptual QPs, denoted as $PQP_G$, $PQP_B$, $PQP_R$, are shown in (21)-(24), respectively:

$$PQP_G = z_G + \left(x_G + \left[6 \times \log_2(g_G)\right]\right) \in \left[\frac{o}{2}, o\right] \tag{21}$$

$$PQP_B = z_{B,R} + \left(x_B + \left[6 \times \log_2(g_B)\right]\right) \in \left[o, o_{\max}\right] \tag{22}$$

$$PQP_R = z_{B,R} + \left(x_R + \left[6 \times \log_2(g_R)\right]\right) \in \left[o, o_{\max}\right] \tag{23}$$

$$o = \frac{1}{w}\sum_{k=1}^{w} p_k = 6 \tag{24}$$

where $z_G$, and $z_{B,R}$ correspond to perceptual QP offsets (temporal masking) via motion vector magnitudes in PUs. Variables $x_G$, $x_B$ and $x_R$, denote the frame-level QPs for the G, B and R channels, respectively. Variables $g_G$, $g_B$ and $g_R$ correspond to the non-normalized variance in each G, B and R CB, respectively. These spatial masking and temporal masking perceptual QP offsets are employed to perceptually adjust the frame-level QP at the CB level (i.e., for each color channel). Variable $o$ refers to the mean CB-level perceptual QP offset in HEVC HM RExt, and constant $o_{\max}$ is the maximum number of CB-level QP offset allowed in HEVC HM RExt ($o_{\max}$ = 12 [29]). Constant $w$ refers to the maximum number of frame-level QP offsets permitted in HEVC HM RExt ($w$ = 12 [29]) and where $p_k$ refers to the $k^{th}$ CB-level QP offset. The perceptual QP offset range is lower for G data, as compared with B and R data, because humans are more sensitive to compression artifacts in G data due to the HVS spectral sensitivity phenomenon (see Figure 1). Variables $z_G$, and $z_{B,R}$ are computed in (25)-(28), respectively.

$$z_G = \begin{cases} \frac{o}{2} & \text{if } D(u,y) > F, \\ 0 & \text{else.} \end{cases} \tag{25}$$

$$z_{B,R} = \begin{cases} o & \text{if } D(u,y) > F, \\ 0 & \text{else.} \end{cases} \tag{26}$$

$$F = \frac{1}{k}\sum_{i=1}^{k} D(u,y) \tag{27}$$

$$D(u,y) = \sqrt{v_x^2 + v_y^2} \tag{28}$$

where $F$ is the arithmetic mean motion vector magnitude in a PU within an entire frame and where $D(u,y)$ corresponds to the magnitude of motion vector $v$ in a PU. More specifically, $D(u,y)$ denotes the magnitude of a motion vector in a PU within the $y^{th}$ CU of the $u^{th}$ frame and where $k$ corresponds to the total number of PUs in the $u^{th}$ frame. Subscripts $x$ and $y$ denote the coordinates $(x,y)$ of motion vector $v$. $D(u,y)$ constitutes an adaptive threshold value related to temporal masking, whereby magnitude $D(u,y)$ of motion vector $v$ must exceed $V$ in order for a region to be considered as high motion. Due to the fact that the G, B and R CBs are of equal size (see Fig. 3), the motion vectors in each G, B and R Prediction Block (PB) do not differ in magnitude (i.e., no scaling is required). This is the reason why Spectral-PQ operates at the PU level in relation to measuring motion vector magnitudes. In terms of perceptual quantization adjustments as a result of temporal masking, recall that the HVS is much less sensitive to compression artifacts in high motion regions of video data [40]. We now define normalized variances and the associated variables in (29)-(37) respectively:

$$a_G = \frac{B_G \cdot g_G + H_G}{g_G + B_G \cdot H_G} \tag{29}$$

$$g_G = 1 + \min\left(\sigma^2_{G_d}\right) \tag{30}$$

$$H_G = \frac{1}{e_G} \sum_{j=1}^{e_G} g_{G_j} \tag{31}$$

$$a_B = \frac{B_B \cdot g_B + H_B}{g_B + B_B \cdot H_B} \tag{32}$$

$$g_B = 1 + \min\left(\sigma^2_{B_d}\right) \tag{33}$$

$$H_B = \frac{1}{e_B} \sum_{j=1}^{e_B} g_{B_j} \tag{34}$$

$$a_R = \frac{B_R \cdot g_R + H_R}{g_R + B_R \cdot H_R} \tag{35}$$

$$g_R = 1 + \min\left(\sigma^2_{R_d}\right) \tag{36}$$

$$H_R = \frac{1}{e_R} \sum_{j=1}^{e_R} g_{R_j} \tag{37}$$

where variable $B$ refers to a scaling factor for normalizing the spatial activity of a G CB ($B = 2$ by default in HEVC HM [29]). Recall that variable $g_G$ in Eq. (37) corresponds to the non-normalized spatial variance of a G CB, $H$ denotes the mean variance of all $2N \times 2N$ G CBs belonging to the current picture of a frame. Variable $\sigma^2_G$ refers to the variance of samples in an $N \times N$ CB sub-block, denoted as $d$ (where $d = 1,....,4$), within a $2N \times 2N$ G CB and $e$ corresponds to the number of $2N \times 2N$ G CBs in the current picture.

**Table 2**: Bitrate reductions achieved by the proposed Spectral-PQ method compared with HM 16.17 and SCM 8.7 in the QP 22 tests on natural content and screen content, respectively.

| Bitrate Reductions (%) and Kbps of Spectral-PQ versus Reference Techniques (%) for QP 22 Tests | | | | | | | |
|---|---|---|---|---|---|---|---|
| **Natural Content** | **Spectral-PQ (SPQ) versus HM** | | | **Screen Content** | **Spectral-PQ (SPQ) versus SCM** | | |
| 30-Bit Sequence | Reduction | SPQ Kbps | HM Kbps | 24-Bit Sequence | Reduction | SPQ Kbps | SCM Kbps |
| **BirdsInCage** | **−80.1%** | 15490.65 | 77841.94 | **BasketballScreen** | **−36.4%** | 13626.50 | 21412.09 |
| **CrowdRun** | **−48.7%** | 128312.14 | 250073.16 | **MissionControlClip** | **−27.6%** | 14645.67 | 20229.62 |
| **DuckAndLegs** | **−66.3%** | 58403.16 | 173273.67 | **CADWaveform** | **−20.2%** | 2508.35 | 3143.54 |
| **Kimono** | **−77.0%** | 11413.39 | 49518.36 | **Desktop** | **−22.2%** | 23906.18 | 30714.04 |
| **OldTownCross** | **−79.6%** | 49411.52 | 242008.03 | **FlyingGraphics** | **−32.8%** | 66723.36 | 99289.63 |
| **ParkScene** | **−73.6%** | 15327.28 | 58088.19 | **PPT_DOC_XLS** | **−24.0%** | 3771.31 | 4964.65 |
| **Seeking** | **−59.2%** | 101464.32 | 248752.70 | **SocialNetworkMap** | **−34.3%** | 148097.30 | 225301.49 |
| **Traffic** | **−57.1%** | 12034.06 | 28029.28 | **VenueVu** | **−36.5%** | 9370.18 | 14756.17 |

**5.0 Evaluation, Results and Discussion**

We implement the proposed technique into the JCT-VC HEVC HM 16.17 RExt + SCM 8.7 codec [29]. Spectral-PQ is compared with two anchors, which are as follows: HEVC HM 16.17 RExt (for the coding of RGB 4:4:4 natural content) and also HEVC SCM 8.7 (for the coding of RGB 4:4:4 screen content); AdaptiveQP is enabled in anchors. The evaluations are conducted on 16 official JCT-VC RGB 4:4:4 video sequences, which are listed in Table 1. The natural content is 10-bits per sample and the screen content is 8-bits per sample; these sequences have a resolution of HD 1080p. In the evaluation, we measure the bitrate reductions Spectral-PQ versus anchors over four QP data points (QPs 22, 27, 32 and 37) using the Random Access (RA) configuration as per the JCT-VC common test conditions specified in [43, 44]. In order to ascertain the perceptual coding efficacy of Spectral-PQ versus anchors, the subjective evaluations — quantified using MOS as per ITU-R Rec. P.910 [45] — and SSIM [46] are the most important perceptual metrics. MOS = 5 equates to imperceptible distortion, MOS = 4 equates to near-imperceptible distortion and MOS = 3 equates to perceptible distortion, but not overly distracting. MOS < 3 usually corresponds to poor subjective quality. As regards the SSIM metric, a score of SSIM > 0.95 is considered to correlate with MOS = 5. SSIM = 1 is the maximum value (i.e., mathematically lossless reconstruction). The subjective tests are conducted on a HD 50 inch display at a viewing distance of 0.75m ≈ 29.5 inch. In line with ITU-R Rec. P.910 [45], four participants engaged in the subjective evaluations; i.e., by analyzing sequences coded using QPs 22, 27, 32 and 37. In the ITU-T P.910 subjective evaluation, the following conditions are recommended:

- Number of participants ≥ 4 and ≤ 40;
- Spatiotemporal analysis for ascertaining perceptual visual quality;
- Compute Mean Opinion Score (MOS).

In comparison with anchors (see Tables 2-5 and Figure 7), Spectral-PQ attains vast bitrate reductions in all tests (i.e., initial QPs 22, 27, 32 and 37). Noteworthy bitrate reductions achieved by Spectral-PQ are as follows: 80.1% (QP 22 test) and 81% (QP 27 test) on the BirdsInCage sequence. In terms of perceptual quality, the subjective evaluation participants chose MOS = 5 in both the QP 22 and QP 27 tests. This means that the Spectral-PQ coded BirdsInCage sequence proved to be visually identical to the BirdsInCage sequence coded by anchors. The SSIM scores in these tests correlated with the subjective score of MOS = 5. In other words, SSIM = 0.9899 (QP 22 test) and SSIM = 0.9892 (QP 27 test). In the QP 37 tests, Spectral-PQ achieves perceptually lossless coding compared with anchors (i.e., HM and SM). Predictably, in the QP 37 tests Spectral-PQ does not achieve visually lossless quality in comparison with the raw data; the same is true for anchors HM and SCM (see Table 11). This is because QP = 37 is a very high initial QP, thus resulting in major quantization-induced compression artifacts in the coded data. Regarding overall MOS and SSIM scores for the 16 sequences shown in Table 1, an MOS = 5 was scored by each participant in 100% of QP 22 tests, 94% of QP 27 tests, 44% of QP 32 tests and 19% of QP 37 tests. An SSIM > 0.95 was recorded in the vast majority of QP 22 and QP 27 tests.

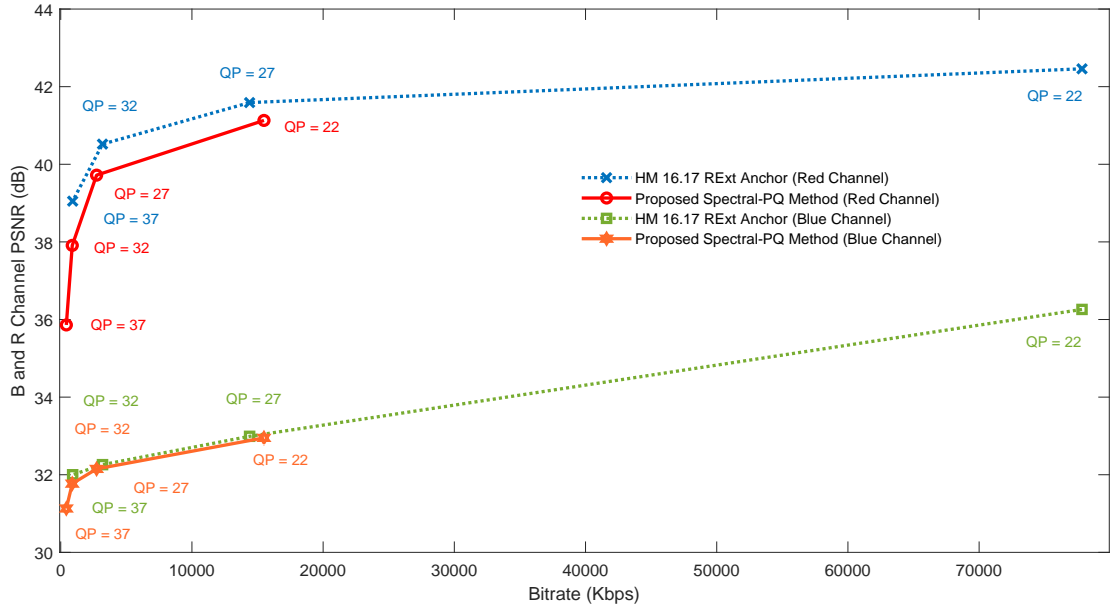

**Figure 7**: A plot showing the bitrate reductions attained by Spectral-PQ (B and R color channels) over four QP data points compared with HM 16.17 RExt on the BirdsInCage RGB 4:4:4 30-bit video sequence (RA configuration).

**Table 3**: Bitrate reductions achieved by the proposed Spectral-PQ method compared with HM 16.17 and SCM 8.7 in the QP 27 tests on natural content and screen content, respectively.

| Bitrate Reductions (%) and Kbps of Spectral-PQ versus Reference Techniques (%) for QP 27 Tests | | | | | | | |
|---|---|---|---|---|---|---|---|
| **Natural Content** | **Spectral-PQ (SPQ) versus HM** | | | **Screen Content** | **Spectral-PQ (SPQ) versus SCM** | | |
| 30-Bit Sequence | Reduction | SPQ Kbps | HM Kbps | 24-Bit Sequence | Reduction | SPQ Kbps | SCM Kbps |
| **BirdsInCage** | **−81.0%** | 2732.45 | 14396.51 | **BasketballScreen** | **−36.6%** | 8797.10 | 13880.52 |
| **CrowdRun** | **−44.2%** | 7218.58 | 12945.72 | **MissionControlClip** | **−30.1%** | 10457.37 | 14965.30 |
| **DuckAndLegs** | **−71.6%** | 13334.18 | 46938.47 | **CADWaveform** | **−22.8%** | 2033.60 | 2634.25 |
| **Kimono** | **−63.1%** | 11413.39 | 49518.36 | **Desktop** | **−26.8%** | 18255.60 | 24948.25 |
| **OldTownCross** | **−80.2%** | 49411.52 | 242008.03 | **FlyingGraphics** | **−34.8%** | 35537.43 | 54541.24 |
| **ParkScene** | **−62.7%** | 5076.82 | 13622.71 | **PPT_DOC_XLS** | **−28.7%** | 2900.02 | 4065.77 |
| **Seeking** | **−53.7%** | 26602.47 | 57433.90 | **SocialNetworkMap** | **−37.2%** | 69549.17 | 110767.32 |
| **Traffic** | **−52.3%** | 4947.75 | 10372.55 | **VenueVu** | **−38.9%** | 4501.67 | 7051.26 |

**Table 4**: Bitrate reductions achieved by the proposed Spectral-PQ method compared with HM 16.17 and SCM 8.7 in the QP 32 tests on natural content and screen content, respectively.

| Bitrate Reductions (%) and Kbps of Spectral-PQ versus Reference Techniques (%) for QP 32 Tests | | | | | | | |
|---|---|---|---|---|---|---|---|
| **Natural Content** | **Spectral-PQ (SPQ) versus HM** | | | **Screen Content** | **Spectral-PQ (SPQ) versus SCM** | | |
| 30-Bit Sequence | Reduction | SPQ Kbps | HM Kbps | 24-Bit Sequence | Reduction | SPQ Kbps | SCM Kbps |
| **BirdsInCage** | **−72.4%** | 878.80 | 3184.89 | **BasketballScreen** | **−39.9%** | 5628.34 | 9368.45 |
| **CrowdRun** | **−42.2%** | 19394.38 | 33567.53 | **MissionControlClip** | **−32.5%** | 7395.05 | 10948.23 |
| **DuckAndLegs** | **−65.4%** | 4669.20 | 13504.09 | **CADWaveform** | **−29.8%** | 1514.38 | 2156.89 |
| **Kimono** | **−50.4%** | 1729.61 | 3489.28 | **Desktop** | **−33.7%** | 13249.18 | 19979.23 |
| **OldTownCross** | **−73.6%** | 2534.30 | 9587.54 | **FlyingGraphics** | **−35.5%** | 20381.24 | 31621.85 |
| **ParkScene** | **−55.6%** | 2102.07 | 4735.86 | **PPT_DOC_XLS** | **−35.7%** | 2037.53 | 3170.20 |
| **Seeking** | **−53.1%** | 8925.13 | 19013.06 | **SocialNetworkMap** | **−41.3%** | 34120.40 | 58133.60 |
| **Traffic** | **−49.6%** | 2360.25 | 4770.99 | **VenueVu** | **−37.7%** | 2281.73 | 3660.90 |

Table 5: Bitrate reductions achieved by the proposed Spectral-PQ method compared with HM 16.17 and SCM 8.7 in the QP 37 tests on natural content and screen content, respectively.

**Bitrate Reductions (%) and Kbps of Spectral-PQ versus Reference Techniques (%) for QP 37 Tests**

| Natural Content | Spectral-PQ (SPQ) versus HM | | | Screen Content | Spectral-PQ (SPQ) versus SCM | | |
|---|---|---|---|---|---|---|---|
| 30-Bit Sequence | Reduction | SPQ Kbps | HM Kbps | 24-Bit Sequence | Reduction | SPQ Kbps | SCM Kbps |
| BirdsInCage | −53.1% | 426.24 | 909.53 | BasketballScreen | −45.8% | 3341.73 | 6169.19 |
| CrowdRun | −42.7% | 8761.74 | 15297.99 | MissionControlClip | −36.7% | 4954.56 | 7827.45 |
| DuckAndLegs | −54.4% | 2106.01 | 4621.24 | CADWaveform | −37.6% | 1057.64 | 1693.81 |
| Kimono | −44.5% | 810.33 | 1459.86 | Desktop | −38.4% | 9007.11 | 14623.50 |
| OldTownCross | −53.5% | 1168.15 | 2512.98 | FlyingGraphics | −35.9% | 11842.58 | 18472.08 |
| ParkScene | −51.6% | 899.44 | 1859.91 | PPT_DOC_XLS | −38.1% | 1322.54 | 2136.77 |
| Seeking | −52.8% | 3577.53 | 7571.58 | SocialNetworkMap | −44.3% | 16201.77 | 29063.32 |
| Traffic | −49.6% | 1177.06 | 2333.61 | VenueVu | −38.9% | 1159.26 | 1898.67 |

Table 6: Reconstruction quality of G, B and R color channels for Spectral-PQ versus the reference techniques in the QP 22 tests. The reconstruction quality per color channel is quantified using the PSNR (dB) metric.

**G, B and R PSNR Values (dB) for Spectral PQ versus Reference Techniques for QP 22 Tests**

| Natural Content | Spectral-PQ (SPQ) versus HM | | | | | | Screen Content | Spectral-PQ (SPQ) versus SCM | | | | | |
|---|---|---|---|---|---|---|---|---|---|---|---|---|---|
| 30-Bit Sequence | G PSNR (dB) | | B PSNR (dB) | | R PSNR (dB) | | 24-Bit Sequence | G PSNR (dB) | | B PSNR (dB) | | R PSNR (dB) | |
| BirdsInCage | 40.54 | 40.67 | 32.95 | 36.26 | 41.13 | 42.46 | BasketballScreen | 45.05 | 46.77 | 39.09 | 45.83 | 39.10 | 46.04 |
| CrowdRun | 34.64 | 35.56 | 31.34 | 35.50 | 31.62 | 35.79 | MissionControlClip | 46.33 | 48.03 | 40.29 | 47.34 | 40.12 | 47.39 |
| DuckAndLegs | 35.94 | 36.77 | 28.89 | 36.04 | 32.95 | 36.76 | CADWaveform | 51.95 | 54.58 | 45.03 | 54.38 | 45.00 | 54.03 |
| Kimono | 40.14 | 40.37 | 32.51 | 35.91 | 37.20 | 39.01 | Desktop | 48.76 | 51.89 | 41.13 | 51.19 | 41.07 | 51.14 |
| OldTownCross | 35.76 | 36.05 | 30.01 | 35.35 | 31.98 | 35.37 | FlyingGraphics | 41.70 | 42.91 | 35.95 | 42.96 | 36.07 | 42.84 |
| ParkScene | 38.26 | 38.59 | 31.90 | 35.76 | 35.01 | 37.83 | PPT_DOC_XLS | 50.18 | 52.84 | 43.36 | 51.75 | 42.81 | 51.68 |
| Seeking | 34.79 | 35.73 | 31.57 | 35.47 | 31.81 | 35.66 | SocialNetworkMap | 37.18 | 37.90 | 32.50 | 37.85 | 32.26 | 37.75 |
| Traffic | 40.72 | 41.25 | 34.71 | 38.17 | 37.49 | 41.40 | VenueVu | 43.16 | 43.31 | 40.49 | 43.34 | 40.31 | 43.30 |

Table 7: Reconstruction quality of G, B and R color channels for Spectral-PQ versus the reference techniques in the QP 27 tests. The reconstruction quality per color channel is quantified using the PSNR (dB) metric.

**G, B and R PSNR Values (dB) for Spectral PQ versus Reference Techniques for QP 27 Tests**

| Natural Content | Spectral-PQ (SPQ) versus HM | | | | | | Screen Content | Spectral-PQ (SPQ) versus SCM | | | | | |
|---|---|---|---|---|---|---|---|---|---|---|---|---|---|
| 30-Bit Sequence | G PSNR (dB) | | B PSNR (dB) | | R PSNR (dB) | | 24-Bit Sequence | G PSNR (dB) | | B PSNR (dB) | | R PSNR (dB) | |
| BirdsInCage | 39.63 | 39.77 | 32.15 | 32.99 | 39.72 | 41.59 | BasketballScreen | 41.16 | 46.77 | 39.09 | 45.83 | 39.10 | 46.04 |
| CrowdRun | 31.01 | 31.28 | 28.66 | 31.80 | 28.86 | 32.16 | MissionControlClip | 42.47 | 44.25 | 36.10 | 43.54 | 35.68 | 43.56 |
| DuckAndLegs | 33.49 | 33.87 | 26.73 | 29.31 | 31.06 | 33.45 | CADWaveform | 47.26 | 50.01 | 39.57 | 49.79 | 39.55 | 49.77 |
| Kimono | 38.43 | 38.72 | 31.67 | 32.66 | 35.25 | 37.30 | Desktop | 43.66 | 46.78 | 35.27 | 46.30 | 35.23 | 46.28 |
| OldTownCross | 34.44 | 34.55 | 29.01 | 30.25 | 31.15 | 32.20 | FlyingGraphics | 36.86 | 37.70 | 31.52 | 37.63 | 31.81 | 37.63 |
| ParkScene | 35.93 | 36.38 | 30.74 | 32.16 | 32.69 | 35.53 | PPT_DOC_XLS | 45.60 | 47.96 | 38.48 | 47.53 | 37.00 | 47.17 |
| Seeking | 31.54 | 31.69 | 29.90 | 31.89 | 29.76 | 32.24 | SocialNetworkMap | 32.87 | 33.45 | 28.47 | 33.34 | 28.10 | 33.11 |
| Traffic | 37.90 | 38.68 | 32.32 | 35.29 | 34.63 | 38.66 | VenueVu | 40.41 | 40.68 | 37.68 | 40.72 | 37.34 | 40.48 |

**Table 8**: Reconstruction quality of G, B and R color channels for Spectral-PQ versus the reference techniques in the QP 32 tests. The reconstruction quality per color channel is quantified using the PSNR (dB) metric.

**G, B and R PSNR Values (dB) for Spectral PQ versus Reference Techniques for QP 32 Tests**

| Natural Content | Spectral-PQ (SPQ) versus HM | | | | | | Screen Content | Spectral-PQ (SPQ) versus SCM | | | | | |
|---|---|---|---|---|---|---|---|---|---|---|---|---|---|
| 30-Bit Sequence | G PSNR (dB) | | B PSNR (dB) | | R PSNR (dB) | | 24-Bit Sequence | G PSNR (dB) | | B PSNR (dB) | | R PSNR (dB) | |
| BirdsInCage | 38.57 | 38.95 | 31.77 | 32.26 | 37.91 | 40.52 | BasketballScreen | 37.21 | 39.14 | 31.23 | 38.47 | 31.10 | 38.50 |
| CrowdRun | 28.57 | 28.86 | 26.38 | 29.43 | 26.58 | 29.71 | MissionControlClip | 38.16 | 40.19 | 31.73 | 39.67 | 31.10 | 39.50 |
| DuckAndLegs | 31.65 | 32.24 | 25.76 | 27.08 | 29.09 | 31.78 | CADWaveform | 42.20 | 45.07 | 34.05 | 45.11 | 34.00 | 44.89 |
| Kimono | 36.38 | 36.89 | 30.83 | 31.77 | 33.20 | 35.50 | Desktop | 38.17 | 41.46 | 30.21 | 41.19 | 30.14 | 41.16 |
| OldTownCross | 33.41 | 33.75 | 28.34 | 29.24 | 30.10 | 31.49 | FlyingGraphics | 32.98 | 33.73 | 27.90 | 33.40 | 28.33 | 33.59 |
| ParkScene | 33.59 | 34.26 | 29.60 | 31.09 | 30.51 | 33.48 | PPT_DOC_XLS | 40.42 | 43.12 | 33.15 | 43.29 | 31.00 | 42.48 |
| Seeking | 30.00 | 30.20 | 28.29 | 30.42 | 27.86 | 30.41 | SocialNetworkMap | 29.41 | 29.93 | 25.30 | 29.74 | 24.81 | 29.40 |
| Traffic | 32.31 | 33.46 | 28.15 | 31.32 | 29.37 | 33.56 | VenueVu | 37.56 | 38.00 | 34.99 | 38.05 | 34.51 | 37.68 |

**Table 9**: Reconstruction quality of G, B and R color channels for Spectral-PQ versus the reference techniques in the QP 37 tests. The reconstruction quality per color channel is quantified using the PSNR (dB) metric.

**G, B and R PSNR Values (dB) for Spectral PQ versus Reference Techniques for QP 37 Tests**

| Natural Content | Spectral-PQ (SPQ) versus HM | | | | | | Screen Content | Spectral-PQ (SPQ) versus SCM | | | | | |
|---|---|---|---|---|---|---|---|---|---|---|---|---|---|
| 30-Bit Sequence | G PSNR (dB) | | B PSNR (dB) | | R PSNR (dB) | | 24-Bit Sequence | G PSNR (dB) | | B PSNR (dB) | | R PSNR (dB) | |
| BirdsInCage | 37.17 | 37.75 | 31.13 | 32.00 | 35.86 | 39.05 | BasketballScreen | 33.22 | 35.17 | 27.35 | 34.67 | 27.18 | 34.59 |
| CrowdRun | 26.24 | 26.64 | 24.26 | 27.14 | 24.51 | 27.42 | MissionControlClip | 33.50 | 35.59 | 27.48 | 35.52 | 26.80 | 35.11 |
| DuckAndLegs | 29.57 | 30.35 | 24.69 | 26.13 | 27.03 | 29.98 | CADWaveform | 36.69 | 39.68 | 28.85 | 39.82 | 28.95 | 39.81 |
| Kimono | 34.05 | 34.76 | 29.74 | 30.98 | 31.12 | 33.46 | Desktop | 32.62 | 35.47 | 25.52 | 35.21 | 25.36 | 35.24 |
| OldTownCross | 31.86 | 32.52 | 27.47 | 28.74 | 28.71 | 30.80 | FlyingGraphics | 29.31 | 30.11 | 24.76 | 29.55 | 25.19 | 29.92 |
| ParkScene | 31.27 | 32.04 | 28.34 | 30.11 | 28.49 | 31.39 | PPT_DOC_XLS | 33.69 | 35.50 | 28.16 | 38.47 | 27.10 | 35.62 |
| Seeking | 28.62 | 28.92 | 26.56 | 28.93 | 25.99 | 28.58 | SocialNetworkMap | 26.20 | 26.65 | 22.67 | 26.52 | 22.09 | 26.06 |
| Traffic | 32.31 | 33.46 | 28.15 | 31.32 | 29.37 | 33.56 | VenueVu | 34.73 | 35.30 | 32.48 | 35.40 | 31.87 | 34.88 |

**Table 10**: Global reconstruction quality of RGB data for Spectral-PQ versus the reference techniques in the QP 22, 27, 32 and 37 tests. The reconstruction quality is quantified using perceptual quality metric SSIM.

**RGB SSIM Values for Spectral PQ versus Reference Techniques for QP 22, 27, 32 and 37 Tests**

| Natural Content | Spectral-PQ (SPQ) versus HM | | | | Screen Content | Spectral-PQ (SPQ) versus SCM | | | |
|---|---|---|---|---|---|---|---|---|---|
| 30-Bit Sequence | QP 22 | QP 27 | QP 32 | QP 37 | 24-Bit Sequence | QP 22 | QP 27 | QP 32 | QP 37 |
| BirdsInCage | 0.9899 | 0.9892 | 0.9893 | 0.9894 | BasketballScreen | 0.9933 | 0.9854 | 0.9681 | 0.9316 |
| CrowdRun | 0.9681 | 0.9517 | 0.9424 | 0.9176 | MissionControlClip | 0.9946 | 0.9891 | 0.9765 | 0.9765 |
| DuckAndLegs | 0.9761 | 0.9288 | 0.8681 | 0.8514 | CADWaveform | 0.9967 | 0.9918 | 0.9785 | 0.9529 |
| Kimono | 0.9449 | 0.9250 | 0.9408 | 0.9334 | Desktop | 0.9971 | 0.9929 | 0.9780 | 0.9476 |
| OldTownCross | 0.9263 | 0.8491 | 0.8349 | 0.8915 | FlyingGraphics | 0.9966 | 0.9923 | 0.9821 | 0.9607 |
| ParkScene | 0.9377 | 0.9015 | 0.9028 | 0.9071 | PPT_DOC_XLS | 0.9946 | 0.9879 | 0.9571 | 0.9390 |
| Seeking | 0.9644 | 0.9397 | 0.9291 | 0.9111 | SocialNetworkMap | 0.9901 | 0.9808 | 0.9594 | 0.9163 |
| Traffic | 0.9726 | 0.9556 | 0.9306 | 0.8956 | VenueVu | 0.9935 | 0.9880 | 0.9756 | 0.9539 |

**Table 11**: The criteria for quantifying the Mean Opinion Score (MOS) with respect to the visual reconstruction quality of a compressed video sequence (compared with the raw video data).

| MOS | Visual Quality Difference |
|---|---|
| 5 | Imperceptible (Visually Lossless) |
| 4 | Very Slightly Perceptible |
| 3 | Moderately Perceptible |
| 2 | Significantly Perceptible |
| 1 | Extremely Obvious |

Table 12: Subjective evaluation Mean Opinion Score (MOS) for Spectral-PQ versus Raw Data for QP 22, 27, 32 and 37 Tests. Note that MOS = 5 indicates perceptually lossless quality (best possible score) and MOS = 1 indicates extremely low quality due to highly visible quantization-induced compression artifacts in the coded data.

| Subjective Evaluation Rounded MOS for Spectral-PQ versus Raw Data for QP 22, 27, 32 and 37 Tests | | | | | | | | | |
|---|---|---|---|---|---|---|---|---|---|
| Natural Content | Spectral-PQ (SPQ) versus Raw Data | | | | Screen Content | Spectral-PQ (SPQ) versus Raw Data | | | |
| 30-Bit Sequence | QP 22 | QP 27 | QP 32 | QP 37 | 24-Bit Sequence | QP 22 | QP 27 | QP 32 | QP 37 |
| BirdsInCage | 5 | 5 | 5 | 4 | BasketballScreen | 5 | 5 | 4 | 4 |
| CrowdRun | 5 | 5 | 4 | 2 | MissionControlClip | 5 | 5 | 4 | 4 |
| DuckAndLegs | 5 | 5 | 4 | 4 | CADWaveform | 5 | 5 | 5 | 5 |
| Kimono | 5 | 5 | 4 | 3 | Desktop | 5 | 5 | 5 | 4 |
| OldTownCross | 5 | 5 | 5 | 4 | FlyingGraphics | 5 | 5 | 5 | 5 |
| ParkScene | 5 | 5 | 4 | 3 | PPT_DOC_XLS | 5 | 5 | 5 | 5 |
| Seeking | 5 | 4 | 4 | 2 | SocialNetworkMap | 5 | 5 | 5 | 4 |
| Traffic | 5 | 5 | 4 | 3 | VenueVu | 5 | 5 | 4 | 4 |

As shown in Tables 6-9, the mathematical reconstruction quality — as quantified by G, B and R PSNR dB scores — is considerably inferior for Spectral-PQ coded sequences in comparison with anchor coded sequences. This is due to the higher levels of perceptual quantization applied to B and R data, in particular, in Spectral-PQ coded data. It is important to note that these substantially inferior G, B and R PSNR dB scores have very little correlation with subjective MOS scores and SSIM scores (except when the initial QP is very high); see Tables 10-12 for confirmation of this. In other words, the PSNR metric is well known to be inadequate in terms of assessing the perceptual quality of an image. SSIM is superior to PSNR in this regard and the human subjective evaluation using MOS remains the gold standard in this sphere of works. The results shown in Table 10 and Table 12 confirm that Spectral-PQ successfully achieves perceptually lossless quality irrespective of the G, B and R PSNR dB scores achieved (as shown in Tables 6-9). In essence, we have provided empirical evidence pertaining to the fact that high levels of perceptual quantization can be applied to G, B and R components without incurring visually conspicuous compression artifacts in the coded sequences. To reiterate, this proved to be the case in all but one of the QP 22 and QP 27 tests (and certain QP 32 tests).

**6.0 Conclusion**

In this paper, we have proposed a perceptual video coding technique known as Spectral Perceptual Quantization (Spectral-PQ). Spectral-PQ exploits the inability of the HVS to discern small gradations between similar colors and also different shades of the same color. This is achieved by virtue of a phenomenon known as HVS spectral sensitivity-related color masking (in addition to spatial masking and temporal masking). For technical convenience, and with application for the HEVC standard, Spectral-PQ operates at the CB level and the PU level in HEVC; perceptual quantization operations take place at the CB level. That is, Spectral-PQ separately adjusts the QSteps in G, B and R CBs according to perceptual color masking at the CU level and motion vector magnitudes at the PU level. In the QP 22 and QP 27 tests, the results show that Spectral-PQ coded sequences are perceptually indistinguishable from the raw, uncompressed sequences. In other words, the Mean Opinion Score (MOS) in the subjective evaluations show that Spectral-PQ successfully achieves perceptually lossless quality compared with HEVC anchors HM 16.17 RExt and SCM 8.7. In other simulations, Spectral-PQ considerably reduces bitrates, with a maximum reduction of approximately 81%, in comparison with anchor 1 (HM 16.17 RExt). In terms of runtimes, there is no difference between Spectral-PQ and HEVC anchors with respect to encoding times and decoding times.